\documentclass[10pt,journal]{IEEEtran}
\IEEEoverridecommandlockouts
\usepackage{cite}
\usepackage{amsmath,amssymb,amsfonts}
\usepackage{algorithmic}
\usepackage{graphicx}
\usepackage{textcomp}
\usepackage{xcolor}
\def\BibTeX{{\rm B\kern-.05em{\sc i\kern-.025em b}\kern-.08em
    T\kern-.1667em\lower.7ex\hbox{E}\kern-.125emX}}

\usepackage{url}
\usepackage{booktabs}
\usepackage{caption}
\usepackage{subfigure}
\usepackage{makecell}  
\usepackage[ruled,linesnumbered]{algorithm2e}
\usepackage{setspace}

\usepackage{amsthm}

\makeatletter
\renewcommand\normalsize{%
\@setfontsize\normalsize\@xpt\@xiipt
\abovedisplayskip 3\p@ \@plus3\p@ \@minus5\p@
\abovedisplayshortskip \z@ \@plus3\p@
\belowdisplayshortskip 6\p@ \@plus3\p@ \@minus3\p@
\belowdisplayskip \abovedisplayskip
\let\@listi\@listI}
\makeatother

\begin{document}

\title{FedLAM: Low-latency Wireless Federated Learning via Layer-wise Adaptive Modulation}
\author{Linping Qu,~\IEEEmembership{Graduate Student Member,~IEEE,} Shenghui Song,~\IEEEmembership{Senior Member,~IEEE,} \\and Chi-Ying Tsui,~\IEEEmembership{Senior Member,~IEEE}
\thanks{Linping Qu, Shenghui Song, and Chi-Ying Tsui are with the Department of Electronic and Computer Engineering, Hong Kong University of Science and Technology, Hong Kong (e-mail: lqu@connect.ust.hk; eeshsong@ust.hk; eetsui@ust.hk).}
}

\maketitle

\begin{abstract}
In wireless federated learning (FL), the clients need to transmit the high-dimensional deep neural network (DNN) parameters through bandwidth-limited channels, which causes the communication latency issue. In this paper, we propose a layer-wise adaptive modulation scheme to save the communication latency. Unlike existing works which assign the same modulation level for all DNN layers, we consider the layers' importance which provides more freedom to save the latency. The proposed scheme can automatically decide the optimal modulation levels for different DNN layers. Experimental results show that the proposed scheme can save up to 73.9\% of communication latency compared with the existing schemes.
\end{abstract}


\section{Introduction}
\label{Introduction}
Federated learning (FL) is a promising learning framework\cite{mcmahan2016federated} where only the deep neural network (DNN) parameters rather than the source data will be uploaded to the server, which enhances the data privacy of users.
However, since large-size DNN parameters are transmitted through bandwidth-limited wireless channels in every communication round and a number of communication rounds are needed, long latency becomes a critical bottleneck for wireless FL.

Existing works tackle the latency issue mainly from two aspects. From the aspect of learning techniques, fast convergence, model quantization and sparsification are proposed. The authors of \cite{wu2021fast, nguyen2020fast, van2020asynchronous} analyzed the convergence rate of FL and proposed methods to reduce the number of communication rounds in FL. The authors of \cite{wang2023communication, luping2019cmfl, li2020ggs} proposed model sparsification to reduce the number of communicated DNN parameters. The authors of \cite{wang2021quantized,lan2023quantization, qu2022feddq} proposed quantization to reduce the bit-width of every DNN parameters. The reduction of communication rounds and parameter size can help reduce the communication latency.
From the aspect of communication techniques, traditional adaptive modulation schemes\cite{reddy2015adaptive,hadi2015adaptive} adjusted the modulation level adaptively, where high modulation level is adopted when the channel has a high signal-to-noise ratio (SNR) and a low modulation level is adopted for a poor-SNR channel. It only considered the channel quality without considering the property of learning. The authors of \cite{xu2021adaptive, xu2023adaptive} proposed new adaptive modulation method which considered both learning property and channel quality, but without considering the importance of different DNN layers and just assigned the same modulation level for all DNN layers, which limits the diversity of assigning adaptive modulation levels.

In this paper, we propose a layer-wise adaptive modulation scheme to reduce the latency of wireless FL. For different training stages and channel quality, not only the DNN adopts a optimized modulation level, but also every layer in that DNN can decide its optimal modulation level, which gets more freedom for modulation levels assignment and enhance the saving ratio of latency.
The contributions of this paper are concluded as follows:\\
\noindent1) We investigate the importance of different DNN layers and quantify their importance by the eigenvalue of the hessian matrix for the given layer.\\
\noindent2) A modified convergence analysis of FL is introduced by taking the layer importance into the analysis. Based on that, an optimization problem is built whose target is to achieve maximum loss drop under given latency.\\
\noindent3) Each DNN layer can decide its own modulation level, which boosts the freedom of modulation assignment and helps to achieve more latency reduction.

The remainder of this paper is organized as follows. In Section~\ref{System Model}, we describe the wireless FL system and give the learning performance model and latency model. In Section~\ref{Problem Formulation and Solution}, we formulate an optimization problem and solve it. In Section~\ref{Experiments and Discussions}, we give our experimental results and discussions. Finally, we conclude our paper in Section~\ref{Conclusion}.

\section{System Model}
\label{System Model}
In this section, we introduce the wireless FL system, and describe the learning performance model and latency model.

\subsection{Wireless Federated Learning}
\label{Wireless Federated Learning}
\begin{figure}[t]
    \begin{center}
    \includegraphics[width=0.5\linewidth]{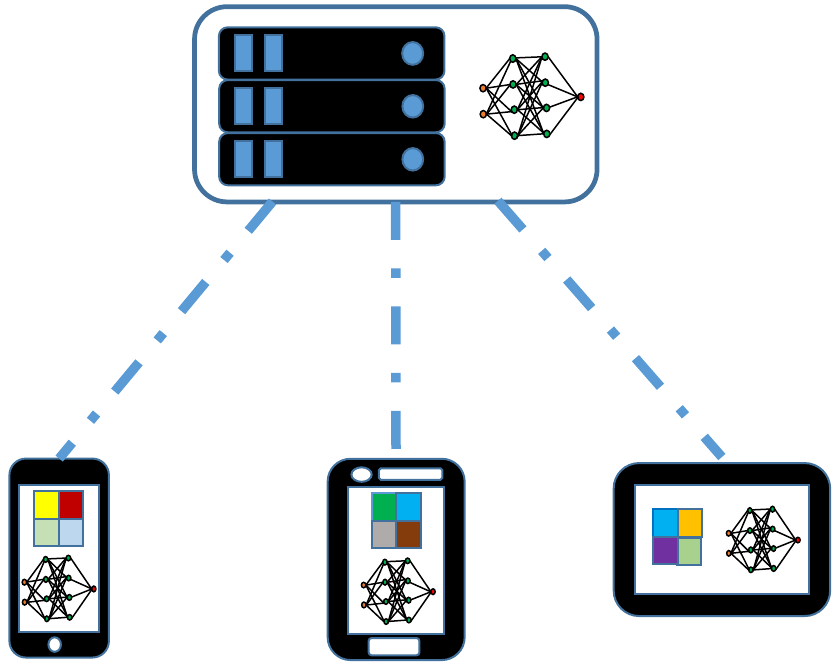}
    \end{center}
    \caption{FL over noisy wireless channels.}
    \label{fig:fl_system}
\end{figure}

As depicted in Fig.~\ref{fig:fl_system}, the FL system contains one remote server and multiple clients.
Basically, FL solves the following optimization problem \cite{li2020federated}:
\begin{align}
{\underset{\mathbf{w}\in {\mathbb{R}^D}}{\rm min} \ f(\mathbf{w}) =\frac{1}{n}\sum_{i=1}^nf_i(\mathbf{w})},
\end{align}
where $\mathbf{w}$ represents the DNN model with $D$ parameters in total, $n$ denotes the number of clients, and $f_i(\mathbf{w})$ denotes the local loss function of the $i$-th client.

In wireless FL, both the uplink and downlink communication channels should be noisy, which introduces communication error to model parameters.
In the $r$-th communication round, the global model at the server, denoted as $\mathbf{w}_r$, will be broadcast to the clients. After receiving the global model, clients will perform $E$ steps of local training based on their local dataset. The local updating process can be modeled as
\begin{align}
\mathbf{w}^i_{r,e+1}=\mathbf{w}^i_{r,e}-\eta{\hat{\nabla}} f_i(\mathbf{w}^i_{r,e}), e=0, \cdots, E-1
\label{receive}
\end{align}
where ${\hat{\nabla}} f_i(\mathbf{w}^i_{r,e})$ represents the stochastic gradient computed from the local mini-batch datasets. And the first step of local training is based on the received noisy global model, which means
\begin{align}
&\mathbf{w}^i_{r,0}=\mathbf{\tilde{w}}_r,
\label{Bw}
\end{align}
where $\mathbf{\tilde{w}}_r$ denotes the model with communication noise.
Upon completing $E$ steps of local training, the $i$-th client obtains the local model update as
\begin{align}
{\Delta \mathbf{w}^i_r=\mathbf{w}^i_{r,E}-\mathbf{w}^i_{r,0}}.
\end{align}
Then, the server aggregates the noisy local model updates by 
\begin{align}
\mathbf{w}_{r+1}=\mathbf{w}_{r}+\frac{1}{n}\sum_{i=1}^n(\tilde{\Delta} \mathbf{w}^i_r).
\end{align}
Above process iterates until FL converges.
\subsection{Learning Performance Model}
Due to the fact that the server can support high transmitting power, in this paper, we assume the downlink has little communication error and only focus on the error in the uplink channels.
For FL with uplink communication bit errors, from \cite{qu2024robust}, the global training loss drop of two adjacent communication rounds can be formulated as
\begin{align}
\begin{split}
\mathbb{E}f(\mathbf{w}_{r})&-\mathbb{E}f(\mathbf{w}_{r+1}) \geq \frac{\eta}{2}\sum_{t=0}^{\tau -1}\mathbb{E}\Vert \nabla f(\bar{\mathbf{w}}_{r,t})\Vert^2\\
&-\frac{L}{2n^2}\sum_{i \in [n]}\mathbb{E}\Vert \tilde{\Delta} \mathbf{w}_r^i-\Delta \mathbf{w}_r^i\Vert^2  \\
&-\frac{L^2(n+1)\tau(\tau-1)\eta^3\sigma^2}{2n} -\frac{L\tau\eta^2\sigma^2}{2n}.
\end{split}
\label{lossdrop_r}
\end{align}
Summing (\ref{lossdrop_r}) over communication round $r=0,...,R-1$, we get the training loss drop for the total learning process:
\begin{align}
\begin{split}
\mathbb{E}f(\mathbf{w}_{0})&-\mathbb{E}f(\mathbf{w}_{R}) \geq \frac{\eta}{2}\sum_{r=0}^{R-1}\sum_{t=0}^{\tau -1}\mathbb{E}\Vert \nabla f(\bar{\mathbf{w}}_{r,t})\Vert^2\\
&-\frac{L}{2n^2}\sum_{r=0}^{R-1}\sum_{i \in [n]}\mathbb{E}\Vert \tilde{\Delta} \mathbf{w}_r^i-\Delta \mathbf{w}_r^i\Vert^2 \\
&-\frac{L^2(n+1)\tau(\tau-1)\eta^3\sigma^2R}{2n} -\frac{L\tau\eta^2\sigma^2R}{2n}.
\end{split}
\label{lossdrop_R}
\end{align}

Above learning convergence assumes every DNN layer the same importance. However, in this paper, we consider the layers' different importance. As widely used in \cite{dong2019hawq,dong2020hawq,yu2022hessian}, the top eigenvalue of the layer's hessian matrix can be used to quantify the DNN layers' importance. Using $H_k$ to denote the importance of the $k$-th layer, (\ref{lossdrop_R}) can be updated into
\begin{align}
\mathbb{E}f(\mathbf{w}_{0})&-\mathbb{E}f(\mathbf{w}_{R}) \geq \frac{\eta}{2}\sum_{r=0}^{R-1}\sum_{t=0}^{\tau -1}\mathbb{E}\Vert \nabla f(\bar{\mathbf{w}}_{r,t})\Vert^2 \nonumber\\
&-\frac{L}{2n^2}\sum_{r=0}^{R-1}\sum_{i \in [n]}\sum_{k \in l}\frac{H_k}{\sum_{k=1}^lH_k}\mathbb{E}\Vert \tilde{\Delta} \mathbf{w}_r^{i,k}-\Delta \mathbf{w}_r^{i,k}\Vert^2 \nonumber\\
&-\frac{L^2(n+1)\tau(\tau-1)\eta^3\sigma^2R}{2n} -\frac{L\tau\eta^2\sigma^2R}{2n}.
\label{lossdrop_R_layer}
\end{align}
The second term on the RHS of (\ref{lossdrop_R_layer}) captures the square error of model updates which is a function of BER as shown in \cite{qu2024energy}. In this work, the BER is impacted by the modulation levels and their relationship can be formulated as \cite{lu1999m}
\begin{align}
b\cong\frac{2}{max(log_2M,2)}\cdot\sum_{i=1}^{max(M/4,1)}Q\left(\sqrt{\frac{2E_s}{N_0}}sin\frac{(2i-1)\pi}{M}\right).
\label{ber_lu}
\end{align}
The authors in \cite{xu2021adaptive} only consider the limited cases of (\ref{ber_lu}), i.e., when $M>4$ and $E_s/N_0\gg1$. But in this paper, we consider the general cases without special requirement on modulation levels and $E_s/N0$.

Based on above relationships, we can get the relationship between learning performance and modulation levels.

\begin{figure*}[htb]
\begin{equation}
\begin{split}
&\underset{M_r^{i,k}}{\rm max}\ \frac{\mathbb{E}f(\mathbf{w}_{0})-\mathbb{E}f(\mathbf{w}_{R})}{T}\\
&=\frac{\frac{\eta}{2}\sum_{r=0}^{R-1}\sum_{t=0}^{\tau -1}\mathbb{E}\Vert \nabla f(\bar{\mathbf{w}}_{r,t})\Vert^2-\frac{L}{2n^2}\sum_{r=0}^{R-1}\sum_{i \in [n]}\sum_{k \in l}\frac{H_k}{\sum_{k=1}^lH_k}\mathbb{E}\Vert \tilde{\Delta} \mathbf{w}_r^{i,k}-\Delta \mathbf{w}_r^{i,k}\Vert^2 -\frac{L^2(n+1)\tau(\tau-1)\eta^3\sigma^2R}{2n} -\frac{L\tau\eta^2\sigma^2R}{2n}}{\frac{DNR}{2B_d}+\frac{VCR}{f}+\sum_{r=0}^{R-1}\sum_{k=1}^l\frac{D_kN}{2B_ulog_2M_r^k}}.
\end{split}
\label{obj_R}
\end{equation}
\end{figure*}

\begin{figure*}[htb]
\begin{equation}
\begin{split}
&\underset{M_r^{i,k}}{\rm max}\ \frac{f(\mathbf{w}_{r})-\mathbb{E}f(\mathbf{w}_{r+1})}{T_r}\\
&=\frac{\frac{\eta}{2}\sum_{t=0}^{\tau -1}\mathbb{E}\Vert \nabla f(\mathbf{w}_{r,t})\Vert^2-\frac{L}{2n}\sum_{k \in l}\frac{H_k}{\sum_{k=1}^lH_k}\mathbb{E}\Vert \tilde{\Delta} \mathbf{w}_r^{i,k}-\Delta \mathbf{w}_r^{i,k}\Vert^2 -\frac{L^2(n+1)\tau(\tau-1)\eta^3\sigma^2}{2n} -\frac{L\tau\eta^2\sigma^2}{2n}}{\frac{DN}{2B_d}+\frac{VC}{f}+\sum_{k=1}^l\frac{D_kN}{2B_ulog_2M_r^k}}.
\end{split}
\label{obj_r}
\end{equation}
\end{figure*}

\subsection{Latency Model}

The latency of wireless FL comprises three parts: downlink broadcasting, local computing, and uplink transmission.
Assuming BPSK is used in downlink broadcasting, the latency can be calculated as
\begin{align}
T_d =\frac{DN}{2B_d},
\label{dn_latency}
\end{align}
where $N$ denotes the bit-length of each DNN parameter, and $B_d$ denotes downlink bandwidth.
Assuming OFDMA technique is used in uplink communication and each client is assigned with a bandwidth of $B_u$, the latency of one client for uplink transmission can be obtained as
\begin{align}
T_u =\sum_{k=1}^l\frac{D_kN}{2B_ulog_2(M^k)},
\label{up_latency}
\end{align}
where $l$ denotes the DNN layers and $D_k$ denotes the parameter size of the $k$-th layer.
And the local computing latency can be calculated as
\begin{align}
T_c=\frac{VC}{f},
\label{compute_latency}
\end{align}
where $V$ represents the volume of the local training data, $C$ denotes the required operation cycles for processing one training data, and $f$ denotes the frequency of the computing unit.

For the entire training which consumes $R$ communication rounds, the total latency is
\begin{align}
T&=\sum_{r=0}^{R-1}T_r=\sum_{r=0}^{R-1}\left(\frac{DN}{2B_d}+\frac{VC}{f}+\sum_{k=1}^l\frac{D_kN}{2B_ulog_2M_r^k}\right) \nonumber\\
&=\frac{DNR}{2B_d}+\frac{VCR}{f}+\sum_{r=0}^{R-1}\sum_{k=1}^l\frac{D_kN}{2B_ulog_2M_r^k}.
\label{latency_allrounds}
\end{align}

\section{Problem Formulation and Solution}
\label{Problem Formulation and Solution}
In this section, we will formulate the latency issue of FL into an optimization problem and then solve it.
\subsection{Problem Formulation}
\label{Problem formulation}
We define the objective function of low-latency FL as (\ref{obj_R}), where the numerator captures the loss drop and the denominator denotes the total time cost of $R$ communication rounds. The meaning of (\ref{obj_R}) is to gain the biggest loss drop at the given time cost. With high modulation level, the time cost in the denominator can be reduced, but the loss drop in the numerator will also be reduced, which shows the trade-off between latency and learning performance.

However, (\ref{obj_R}) is just an ideal objective function because whose solution relies on the overall information of all clients and of all communication rounds. In practice, each client needs to make real-time decision without knowing mutual and future information. So we modify (\ref{obj_R}) into (\ref{obj_r}) to make it practical, which only requires the information for a given client in the current communication round.

\subsection{Problem Solution}
First, to quantify the DNN layers' importance, the power iteration method \cite{yao2020pyhessian} can be used to simplify the calculation of top eigenvalue.

Then, given the objective function of (\ref{obj_r}), by choosing different modulation levels for different DNN layers, we can get different value of (\ref{obj_r}). Considering the fact that the modulation levels are discrete integers, like [2, 4, 8, 16]-PSK for example, we can find the optimal solution of (\ref{obj_r}) by enumeration.

For $l$- layers DNN model, each layer can be assigned with the provided 4 different PSK modulation schemes and the searching space is $4^l$. When $l$ is large, it will take noticeable time to find the optimal solution. To tackle this issue, we further propose a group method for very deep-layers DNN. The layers having similar importance score will be assigned with a same modulation level. Take the 20-layers DNN for example, the original searching space is $4^{20}$ which is too huge to find the optimal solution. But if the layers are divided into 5 groups based on their importance, the searching space will be reduced into $4^5$ which is easy to solve.

\section{Experiments and Discussions}
\label{Experiments and Discussions}
In this section, we conduct experiments to verify the performance of the proposed layer-wise adaptive modulation scheme.
\subsection{Experiment Settings}
\label{Experiment Settings}

The communication settings are similar to the ``AM" \cite{xu2023adaptive}. And we considered three sets of FL tasks: MNIST\cite{lecun1998mnist} dataset classified by LeNet-300-100\cite{han2015deep},
Fashion-MNIST\cite{xiao2017fashion} dataset classified by vanilla CNN \cite{mcmahan2017communication},
CIFAR-10\cite{Krizhevsky09learningmultiple} dataset classified by ResNet-20\cite{he2016deep}.
The datasets were split in an identical and independently distributed (i.i.d) manner. The local training was configured with a learning rate of 0.01, using the SGD optimizer. The number of local training steps was set to 5.

\subsection{ Results and Analysis}
\label{ Results and Analysis}

\begin{figure}[t]
\small
\begin{minipage}[b]{.48\linewidth}
  \centering
  \centerline{\includegraphics[width=4.0cm]{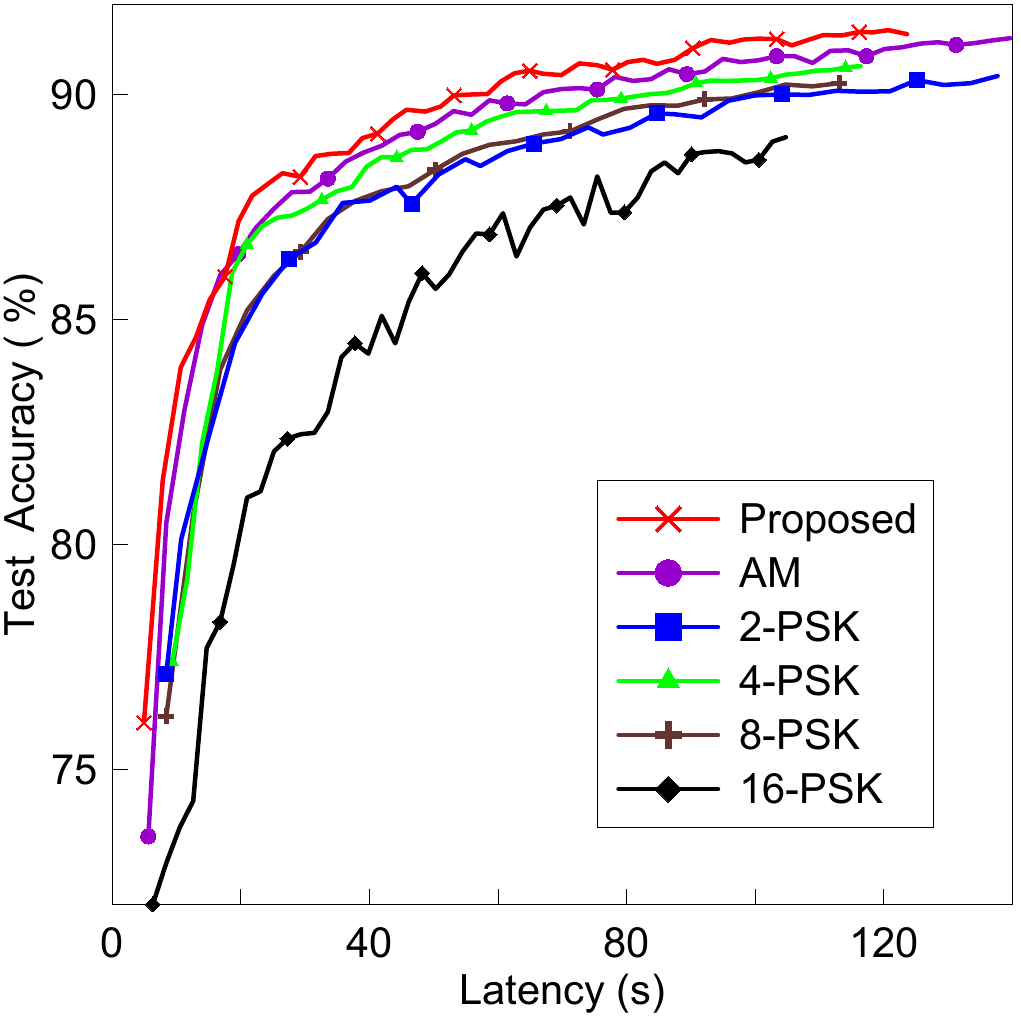}}
  \begin{center}
  (a) Test accuracy vs. communication latency.
  \end{center}
\end{minipage}
\hfill
\begin{minipage}[b]{0.48\linewidth}
  \centering
  \centerline{\includegraphics[width=4.0cm]{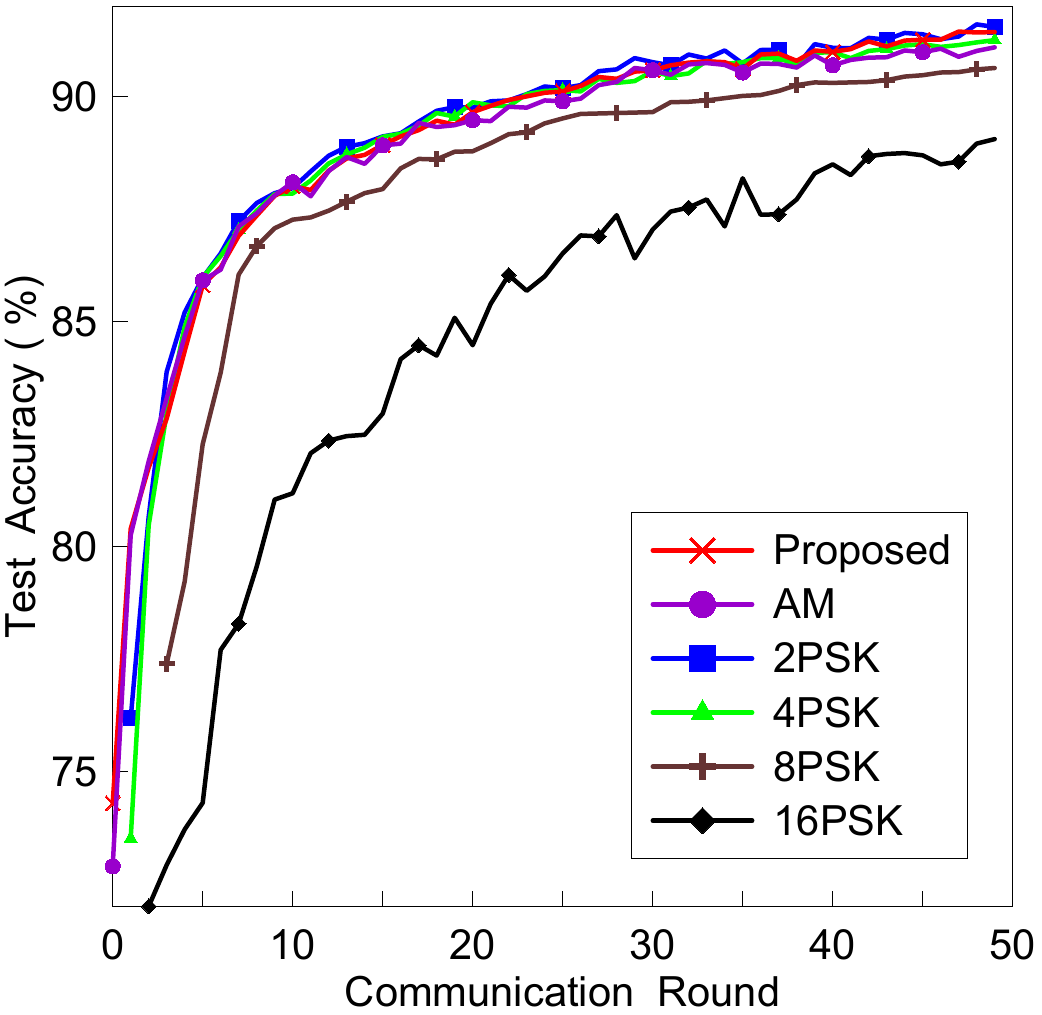}}
  \begin{center}
  (b) Test accuracy vs. communication round.
  \end{center}
\end{minipage}
\hfill
\begin{minipage}[b]{0.48\linewidth}
  \centering
  \centerline{\includegraphics[width=4.0cm]{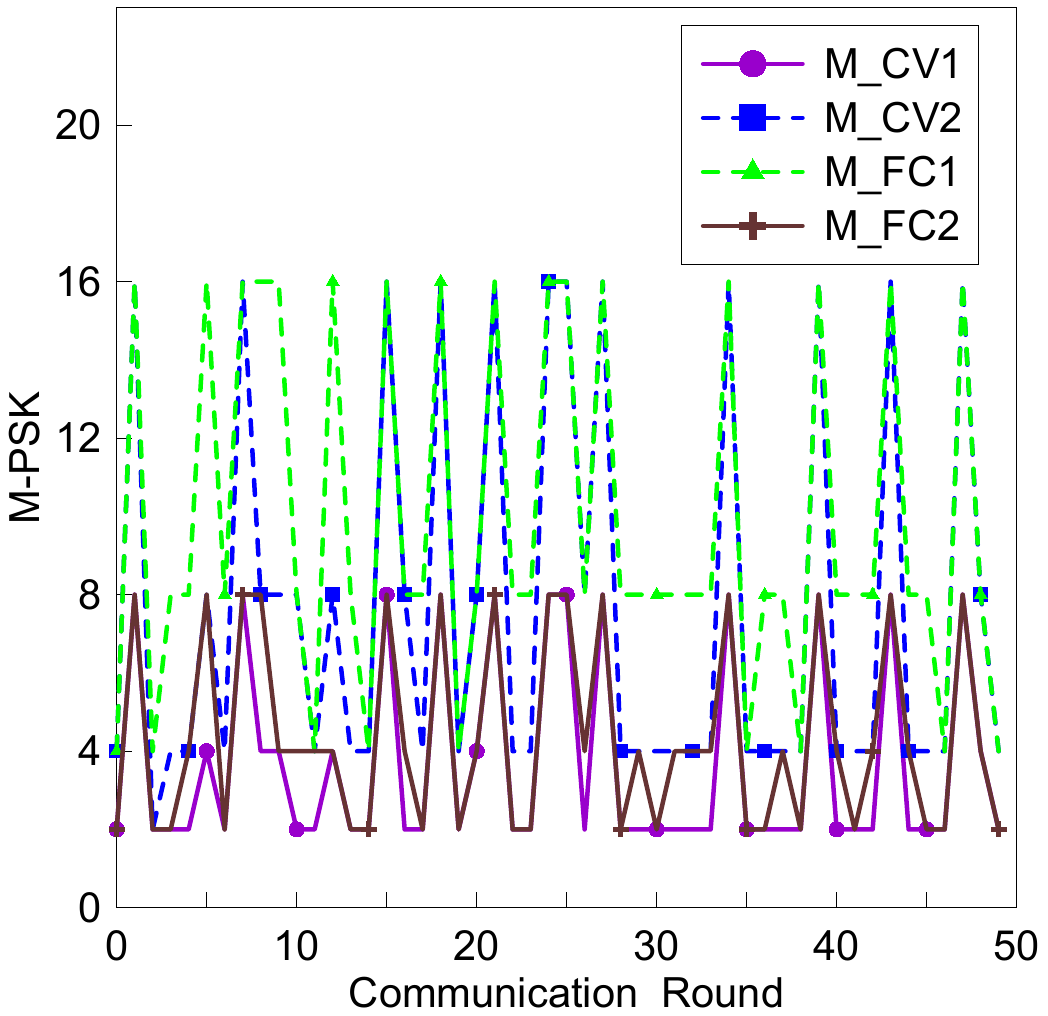}}
  \begin{center}
  (c) Modulation level of each layer changes in training.
  \end{center}
\end{minipage}
\hfill
\begin{minipage}[b]{0.48\linewidth}
  \centering
  \centerline{\includegraphics[width=4.0cm]{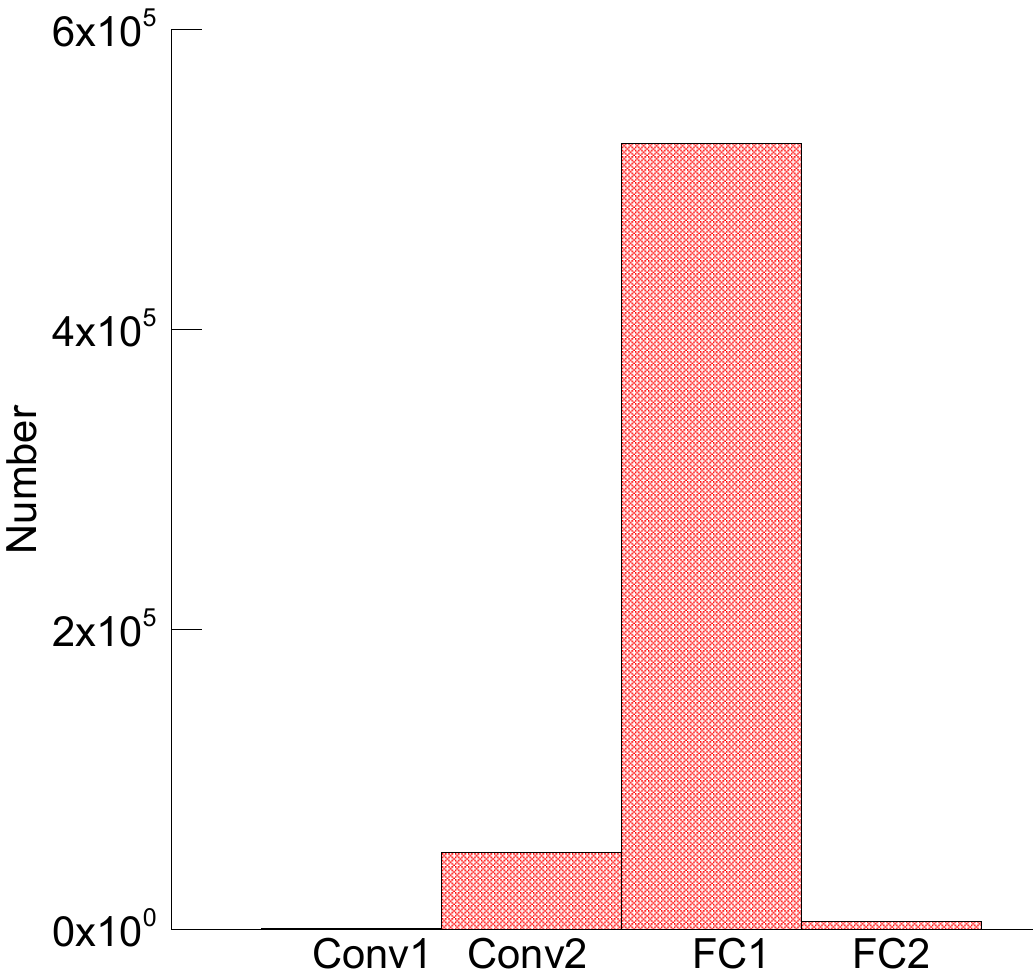}}
  \begin{center}
  (d) Parameter size of each layer in the DNN model.
  \end{center}
\end{minipage}
\caption{Experiment of Fashion-MNIST.}
\label{fig:fashion}
\end{figure}

\begin{figure}[t]
\small
\begin{minipage}[b]{.48\linewidth}
  \centering
  \centerline{\includegraphics[width=4.0cm]{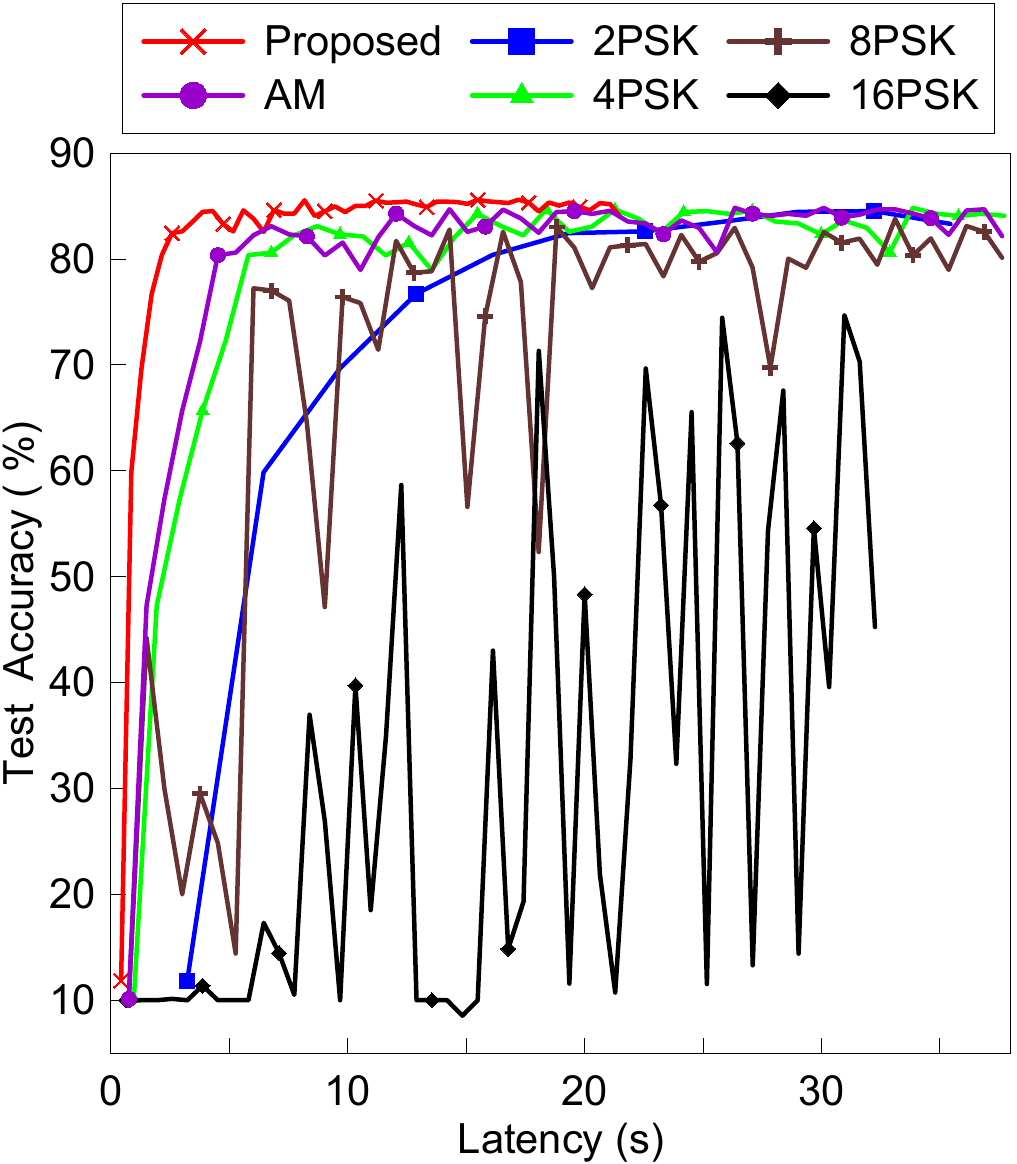}}
  \begin{center}
  (a) Test accuracy vs. communication latency.
  \end{center}
\end{minipage}
\hfill
\begin{minipage}[b]{0.48\linewidth}
  \centering
  \centerline{\includegraphics[width=4.0cm]{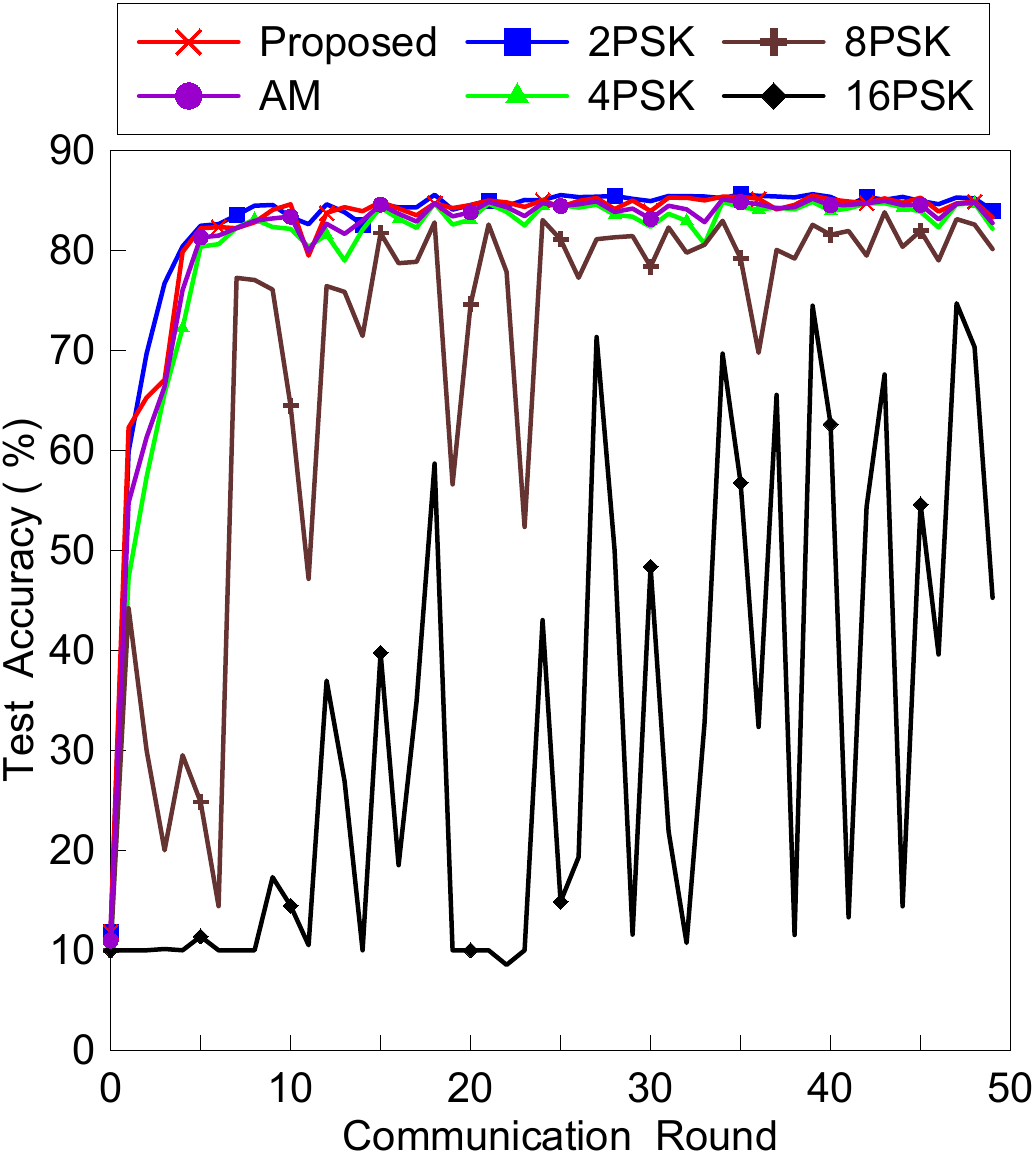}}
  \begin{center}
  (b) Test accuracy vs. communication round.
  \end{center}
\end{minipage}
\hfill
\begin{minipage}[b]{0.48\linewidth}
  \centering
  \centerline{\includegraphics[width=4.0cm]{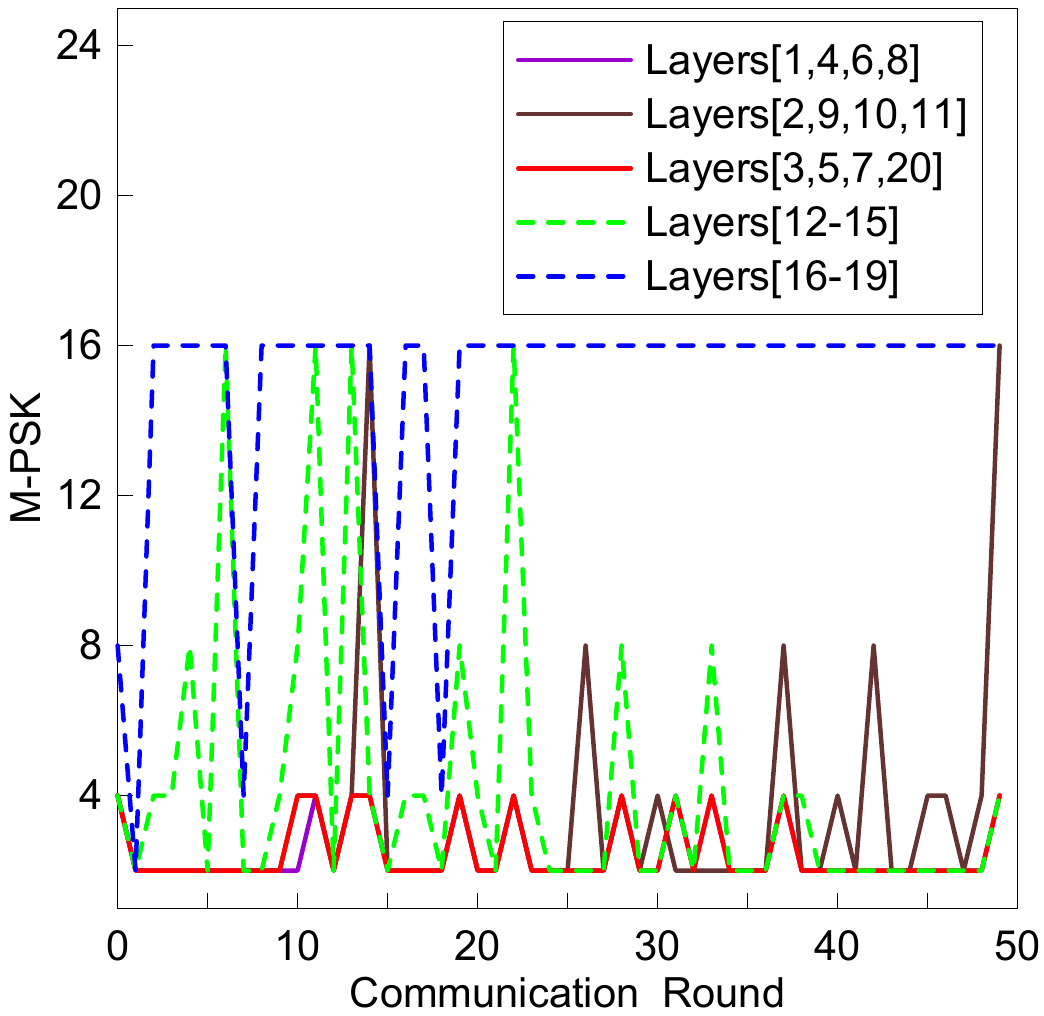}}
  \begin{center}
  (c) Modulation level of each layer changes in training.
  \end{center}
\end{minipage}
\hfill
\begin{minipage}[b]{0.48\linewidth}
  \centering
  \centerline{\includegraphics[width=4.0cm]{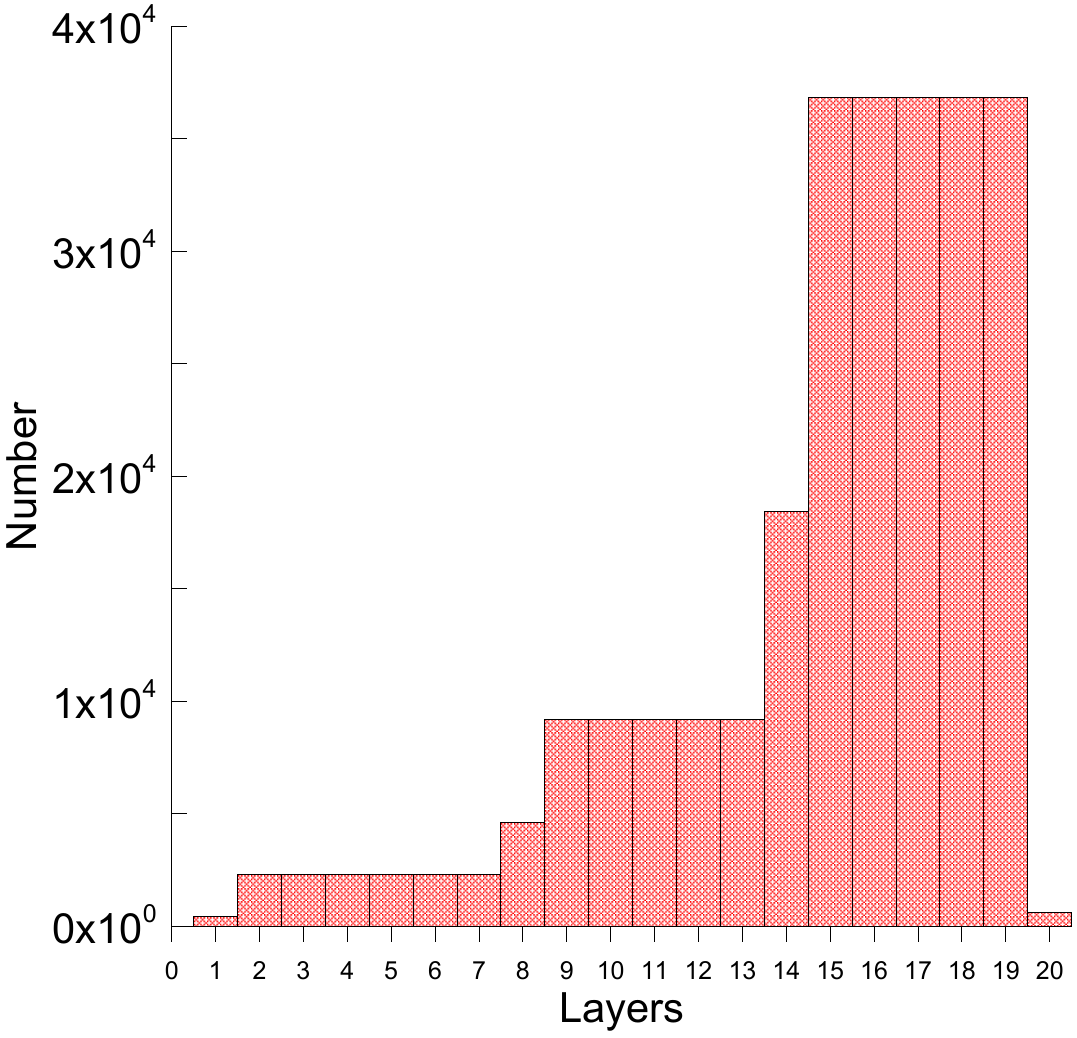}}
  \begin{center}
  (d) Parameter size of each layer in the DNN model.
  \end{center}
\end{minipage}
\caption{Experiment of CIFAR-10.}
\label{fig:cifar}
\end{figure}

Fig.~\ref{fig:fashion} shows the experimental results of Fashion-MNIST. The results of MNIST is similar, so we omit its figures due to page limit. Fig.~\ref{fig:fashion}(a) shows that the proposed scheme consumes least latency for achieving the same test accuracy since it has more freedom to change the modulation for each layer. Fig.~\ref{fig:fashion}(b) shows how learning accuracy improves with communication rounds.
Fig.~\ref{fig:fashion} (c) shows how the modulation levels of every layer changes during training. It can be observed that the first and last layers (CV1 and FC2) generally adopt higher modulation levels than the middle layers (CV2, FC1). This can be explained that the first and last layer are the more important layers which determine the input feature and output result. So low modulation levels are preferred to avoid errors and guarantee the correctness of input and output. And from Fig.~\ref{fig:fashion}(d), the parameter size of the first or the last layer is very small compared with the middle two layers, so the low modulation levels in the first and last layer introduces negligible latency. 

In the CIFAR-10 experiment as shown in Fig.~\ref{fig:cifar}, the layers for ResNet-20 is 20. So the layers group method is adopted to divide 20 layers into 5 groups. The first eight layers and the last layer are assigned with low modulation levels due to their importance and small size. The middle layers, especially for the [16-19] layers that have large parameter size are assigned with high modulation levels for saving latency purpose. 

Concluding all the three experiments, the first and the last layer with higher importance and smaller parameter size are preferred to be assigned with low modulation levels. The middle layers generally with less importance and larger parameter size are suggested to assign higher modulation levels.

The detailed data for performance comparison is summarized in Table.~\ref{tab:summary}. The comparison is based on the latency for achieving a given test accuracy. Since the test accuracy of ``16PSK" is too bad, there is no need to compare with it and thus not listed in the table. The proposed scheme is compared with ``2PSK", ``4PSK", ``8PSK", and ``AM". The saving ratio is for the proposed scheme comparing with the ``AM" scheme.

\begin{table}[t]
\caption{Summary of Experimental Results.}
\centering
\begin{tabular}{|c|c|c|c|}
\hline  
&\makecell[c]{MNIST\\ Acc.=97.8\%}&\makecell[c]{Fashion-MNIST\\ Acc.=90.6\%}&\makecell[c]{CIFAR-10\\ Acc.=83.1\%}\\
\hline  
\makecell[c]{2PSK}&46.4s&\makecell[c]{137.7s}&\makecell[c]{25.8s}\\
\hline  
\makecell[c]{4PSK}&39.3s&\makecell[c]{114.1s}&\makecell[c]{21.3s}\\
\hline  
\makecell[c]{8PSK}&37.3s&\makecell[c]{117.3s}&\makecell[c]{36.1s}\\
\hline
\makecell[c]{AM\cite{xu2023adaptive}}&28.8s&\makecell[c]{94.9s}&\makecell[c]{14.3s}\\
\hline
\makecell[c]{Proposed}&21.6s&\makecell[c]{72.6s}&\makecell[c]{3.5s}\\
\hline
\makecell[c]{Saving}&25\%&\makecell[c]{23.5\%}&\makecell[c]{73.9\%}\\
\hline
\end{tabular}
\label{tab:summary}
\end{table}

\subsection{The Cost of Computing Hessian Eigenvalue}
The proposed layer-wise modulation scheme needs to calculate the hessian matrix of every layer, which introduces extra latency. The hessian value calculation is equivalent to one further step of differential on gradients\cite{yan2022swim}. So the extra overhead of hessian calculation is same as one extra step of back propagation. Due to the advanced AI hardware like GPU and AI chips, the latency bottleneck is more on communication side rather than computation side\cite{yao2018two}. So the one step of back propagation will not introduce noticeable latency compared with the overall latency.

\section{Conclusion}
\label{Conclusion}
In this paper, we investigated the latency issue of wireless FL. By theoretically analyzing the impact of modulation on learning performance and latency, we formulated an optimization problem and proposed the DNN layer-wise adaptive modulation scheme. The proposed scheme can adjust the uplink modulation levels for every DNN layer individually, which can achieve a good trade-off between learning performance and latency.
\clearpage
\bibliographystyle{IEEEbib}
\bibliography{strings,refs}

\end{document}